\documentclass{article}
\usepackage[utf8]{inputenc}
\usepackage[T1]{fontenc}
\usepackage{hyperref}
\usepackage{amsmath}
\usepackage{amsfonts}
\usepackage{nicefrac}
\usepackage{microtype}
\usepackage{natbib}
\usepackage{doi}
\usepackage{geometry}
\title{The Algebra of Meaning: Why Machines Need Montague More Than Moore's Law}

\author{
  Cheonkam Jeong\thanks{Corresponding author. Contact: \texttt{cheonkamjeong@savassan.com}} \\
  Savassan \\
  \texttt{cheonkamjeong@savassan.com}
  \and
  Sungdo Kim\\
  Savassan \\
  \texttt{sdk@savassan.com}
  \and
  Jewoo Park \\
  Savassan \\
  \texttt{jewoop@savassan.com}
}
\date{}

\begin{document}

\maketitle

\begin{abstract}
\noindent Contemporary language models are fluent yet routinely mis-handle the \emph{types} of meaning their outputs entail. We argue that hallucination, brittle moderation, and opaque compliance outcomes are symptoms of missing type-theoretic semantics rather than data or scale limitations. Building on Montague’s view of language as typed, compositional algebra, we recast alignment as a parsing problem: natural-language inputs must be compiled into structures that make explicit their descriptive, normative, and legal dimensions under context.

We present \textit{Savassan}, a neuro-symbolic architecture that compiles utterances into Montague-style logical forms and maps them to typed ontologies extended with deontic operators and jurisdictional contexts. Neural components extract candidate structures from unstructured inputs; symbolic components perform type checking, constraint reasoning, and cross-jurisdiction mapping to produce compliance-aware \emph{guidance} rather than binary censorship. In cross-border scenarios, the system “parses once” (e.g., \texttt{defect\_claim(product\_x, company\_y)}) and projects the result into multiple legal ontologies (e.g., defamation risk in KR/JP, protected opinion in US, GDPR checks in EU), composing outcomes into a single, explainable decision.

This paper contributes: (i) a diagnosis of hallucination as a type error; (ii) a formal Montague--ontology bridge for business/legal reasoning; and (iii) a production-oriented design that embeds typed interfaces across the pipeline. We outline an evaluation plan using legal reasoning benchmarks and synthetic multi-jurisdiction suites. Our position is that trustworthy autonomy requires compositional typing of meaning, enabling systems to reason about what is described, what is prescribed, and what incurs liability within a unified algebra of meaning.
\end{abstract}

\section*{The Paradox}

We've engineered machines capable of crafting poetry, yet they lack the judgment to discern its ethical implications. They draft intricate legal contracts but routinely hallucinate fundamental laws \citep{fournier2024benchmark}. They moderate content at hyperspeed, offering decisions without explanation. This isn't a failure of engineering; it's a profound failure of imagination. We persist in teaching AI to `think' when our true ambition should be to teach it to parse meaning itself.

Consider this: When a user posts ``The CEO is a fraud'' on your platform, conventional AI identifies patterns such as negative sentiment, a named entity, and potential risk. But it fundamentally misses that `fraud' is simultaneously a legal claim demanding evidence, a moral judgment implying intent, and a speech act that directly creates liability \citep{jiao2024ethics}. It merely matches patterns, failing to grasp that this single sentence operates across multiple semantic dimensions, each with distinct truth conditions and critical consequences.

\section*{The Wittgensteinian Trap}

Today's Large Language Models (LLMs) are ensnared in precisely the predicament Wittgenstein warned us about: they're playing language games without truly knowing the rules. They are semantic savants confined within syntactic cages. GPT-4 might pass the bar exam, but it cannot articulate why classifying someone as a ``public figure'' fundamentally alters the legal semantics of defamation. It knows the words, but not their types.

This is precisely where current approaches falter, defaulting to throwing more data at the problem. Bigger models. Better benchmarks. Reinforcement learning from human feedback. But what if we're optimizing for the wrong outcome? What if hallucination isn't a bug, but an inherent feature---the natural consequence of intelligence devoid of ontology? \citep{kampik2023large,ming2024reality} 

\section*{Enter Montague}
In the 1970s, Richard Montague introduced a groundbreaking formal system for representing the semantics of natural language. He demonstrated that linguistic meaning could be encoded using the tools of formal logic, specifically, typed lambda calculus, where meanings compose recursively with mathematical precision \citep{montague1973ptq,hirst1983foundation,retore2013montague}. Far from metaphor, this approach treated language as algebraic in structure, offering a framework where natural language sentences could be parsed, transformed, and interpreted within a rigorously defined type-theoretical system.

Yet, Montague built more than he knew. He crafted the very blueprint for machines capable of navigating the intricate space between pattern and principle. When he wrote that ``every lawyer'' maps to \(\lambda P.\lambda Q.\forall x[lawyer(x) \rightarrow Q(x)]\), he wasn't just performing linguistics. He was revealing how to construct AI that discerns the critical difference between ``every lawyer'' and ``any lawyer''---a distinction that can fundamentally alter liability, scope, and obligation \citep{halvorsen1986montague}.

\section*{The Categorical Imperative for Machines}

Kant implored us to act only according to maxims we could will to be universal laws. But here's a crucial twist---machines already operate on universals. Every neural network, in essence, asks: ``What if everyone who observed this pattern behaved this way?'' \citep{jastroch2020trusted}

The challenge isn't that machines inherently lack ethics; it's that they lack types. They cannot distinguish between:
\begin{itemize}
  \item Descriptive universals: (``All swans are white'')
  \item Normative universals: (``All persons deserve dignity'')
  \item Logical universals: (``All contradictions are false'')
\end{itemize}

Montague Grammar provides an unprecedented solution: a method to encode the categorical imperative, not as a predefined rule, but as a robust type system \citep{retore2013montague}. When our system encounters ``hate speech,'' it doesn't just pattern-match against prohibited words. It parses the semantic structure to understand: this utterance treats a particular as a universal, thereby violating the fundamental type constraint of human dignity.

\section*{Building Savassan: From Philosophy to Production}

Savassan isn't merely another chatbot or content filter. It is the first production-ready system to treat meaning as a compositional algebra \citep{feldstein2024mapping,lu2024survey,pulicharla2025neurosymbolic}. Here's how it manifests in practice:

\subsection*{Real-World Scenario: Cross-Border Content Compliance}

Imagine a Korean user posting about a Japanese company's product defect. The same content simultaneously faces:
\begin{itemize}
  \item Korean law: Strict defamation rules, where truth is not always an absolute defense.
  \item Japanese law: Emphasizing corporate reputation protection.
  \item US law: Section 230 immunity, with a high bar for defamation.
  \item EU law: GDPR implications if personal data is involved.
\end{itemize}

Traditional approach: Run four separate classifiers, then apply the most restrictive outcome.

Savassan's approach:
\begin{itemize}
  \item Parse once: \texttt{defect\_claim(product\_x, company\_y)}
  \item Map to multiple legal ontologies:
    \begin{itemize}
      \item Korea $\rightarrow$ \texttt{potential\_criminal\_defamation}
      \item Japan $\rightarrow$ \texttt{commercial\_disparagement}
      \item US $\rightarrow$ \texttt{protected\_opinion}
      \item EU $\rightarrow$ \texttt{check\_personal\_data\_exposure}
    \end{itemize}
  \item Compose: $\otimes (kr\_risk, jp\_risk, us\_protection, eu\_compliance)$
\end{itemize}

Result: Semantic guidance, not binary censorship.

\subsection*{Neural Pattern Extraction from Unstructured Inputs}

Savassan's architecture begins at the unstructured frontier—free-form user utterances, platform logs, or legal text fragments. A neural preprocessor, based on transformer encoders fine-tuned for semantic anomaly detection, extracts latent patterns such as emerging slang, policy circumvention tactics, or shifts in discourse modality.

However, rather than applying these patterns directly, Savassan subjects each candidate to symbolic validation. Extracted patterns are matched against legal or business ontologies and filtered through type-checkers before activation. These ontologies implement a layered type system: a core level comprising \texttt{Entity}, \texttt{Event}, \texttt{Property}, and \texttt{Relation}, and domain-specific extensions such as legal \texttt{Constraint} types or contextual predicates under \texttt{Context}. This layered structure allows Savassan to generalize semantic parsing across domains while enforcing domain-specific rules where appropriate.

In this sense, Savassan enacts a bottom-up learning cycle constrained by top-down types: neural layers discover, symbolic layers verify. This general pattern (i.e., bottom-up neural extraction with top-down symbolic validation) has precedent in neuro-symbolic literature \citep{feldstein2024mapping,lu2024survey}. However, most existing systems either treat symbolic modules as post-hoc filters or lack type-level enforcement across domain ontologies.

Savassan differs in that it embeds typed interfaces across the entire pipeline: every extracted structure must conform to formal legal or business type systems before propagation. This tight coupling enables jurisdiction-specific alignment and provable constraints on downstream actions.

\section*{The Alignment Problem is a Parsing Problem}

The AI community is preoccupied with alignment \citep{rodemann2025alignment,yi2023alignment}. But alignment to what? The profound insight here is this: alignment isn't about imbuing machines with our values; it's about equipping them to understand how values compose \citep{rossi2019bounded}.

When a platform must balance free speech against harm prevention, it isn't choosing sides. It is resolving a type conflict:
\begin{align*}
  &\texttt{free\_speech}: \langle \text{utterance}, \text{protected} \rangle \\
  &\texttt{harm\_prevention}: \langle \text{utterance}, \text{restricted} \rangle \\
  &\texttt{resolution}: \lambda u.\text{context}(u) \rightarrow \text{precedence}(\text{type1}, \text{type2})
\end{align*}

\section*{The Graph is Not the Territory}
Our legal knowledge graph is far more than a database; it's a navigable semantic space. Each node is a typed concept imbued with compositional rules \citep{bein2025kraft,fan2016process,thomas2009semantic}. Each edge is a semantic transformation. RL agents reason through the graph, discovering not the shortest but the semantically valid path. Savassan’s ontologies support compositional traversal via typed relations, enabling each semantic edge to enforce domain-level constraints during reasoning.

\section*{Why This Matters Now}

We are rapidly entering an era where:
\begin{itemize}
  \item Every platform is effectively a publisher.
  \item Every utterance carries potential liability.
  \item Every algorithmic decision demands explanation \citep{mouta2024ethics}.
  \item Every market is global, governed by local rules.
\end{itemize}

\section*{The Uncomfortable Truth}
Much of today's AI development proceeds without a shared definition of intelligence, without formal semantics for alignment, and without a coherent theory of meaning for content moderation. Montague Grammar is more than a linguistic framework—it is a proof of concept that language can be modeled with logical precision. Savassan builds on this tradition, asserting that AI must be formally grounded in its understanding of what it means to be human.

\section*{The Challenge and the Vision}
The central challenge is not to generate more plausible sentences, but to interpret what those sentences mean in context. A single utterance may assert a fact, express a normative judgment, or trigger a legal obligation. These distinctions are not stylistic; they are semantic, and they have consequences. To meet this challenge, we need AI systems that do not merely approximate language, but model its internal structure—systems that can distinguish between what is described and what is prescribed, between facts and norms, and between statements that inform and those that commit.

Savassan is built on this principle. It treats meaning as structured, typed, and composable. Rather than relying solely on statistical fluency, it analyzes each input for its semantic implications within legal, ethical, and institutional domains. As Wittgenstein observed, ``the meaning of a word is its use in the language'' \citep{wittgenstein1953investigations}. Savassan extends this insight into engineering practice: to align machines with human values, we must first align them with the semantic structures that govern human communication.

\bibliographystyle{plainnat}
\bibliography{references}

\end{document}